\documentclass{article}
\usepackage{arxiv/arxiv}

\usepackage{url}
\usepackage{epsfig}
\usepackage{graphicx}
\usepackage{amsmath}
\usepackage{amssymb}
\usepackage{cancel}

\usepackage{float}
\usepackage{enumitem}
\usepackage{gensymb}
\usepackage{stmaryrd}
\usepackage{fancyhdr}
\usepackage{makecell}
\usepackage{bm}
\usepackage{bbm} 
\usepackage{relsize}

\title{Variational Saccading: Efficient Inference for Large Resolution
  Images}

\newcommand*\samethanks[1][\value{footnote}]{\footnotemark[#1]}

\author{Jason Ramapuram\thanks{University of Geneva}\ \ \thanks{Geneva School of
    Business Administration, HES-SO} \\
  {\tt\small Jason.Ramapuram@etu.unige.ch} \\
  \And
  Maurits Diephuis\samethanks[1] \\
  {\tt\small Maurits.Diephuis@unige.ch}
  \And
  Frantzeska Lavda\samethanks[1]\ \ \samethanks[2] \\
  {\tt\small frantzeska.lavda@hesge.ch}
  \AND
  Russ Webb\thanks{Apple Inc} \\
  {\tt\small rwebb@apple.com}
  \And
  Alexandros Kalousis\samethanks[1]\ \ \samethanks[2] \\
  {\tt\small alexandros.kalousis@hesge.ch}
}

\begin{document}

\maketitle

\begin{abstract}
  Image classification with deep neural networks is typically
restricted to images of small dimensionality such as
$\mathbb{R}^{224\times244}$ in Resnet models \cite{he2016deep}. This
limitation excludes the $\mathbb{R}^{4000\times3000}$ dimensional
images that are taken by modern smartphone cameras and smart devices.
In this work, we aim to mitigate the prohibitive inferential and
memory costs of operating in such large dimensional spaces.
To sample from the high-resolution original input distribution, we
propose using a smaller proxy distribution to learn the
co-ordinates that correspond to regions of interest in the 
high-dimensional space.  We introduce a new principled variational
lower bound that captures the relationship of the proxy distribution's
posterior and the original image's co-ordinate space in a way that
maximizes the conditional classification likelihood. We empirically
demonstrate on one synthetic benchmark and one real world large
resolution DSLR camera image dataset that our method produces
comparable results with $\sim$\textbf{10x} faster inference and lower memory consumption than a
model that utilizes the entire original input distribution. Finally,
we experiment with a more complex setting using
mini-maps from Starcraft II \cite{vinyals2017starcraft} to infer
the number of characters in a complex 3d-rendered scene. Even in such
complicated scenes our model provides strong localization: a feature
missing from traditional classification models.
\end{abstract}

\section{Introduction}

\begin{figure}
  \begin{center}
    \includegraphics[width=\linewidth]{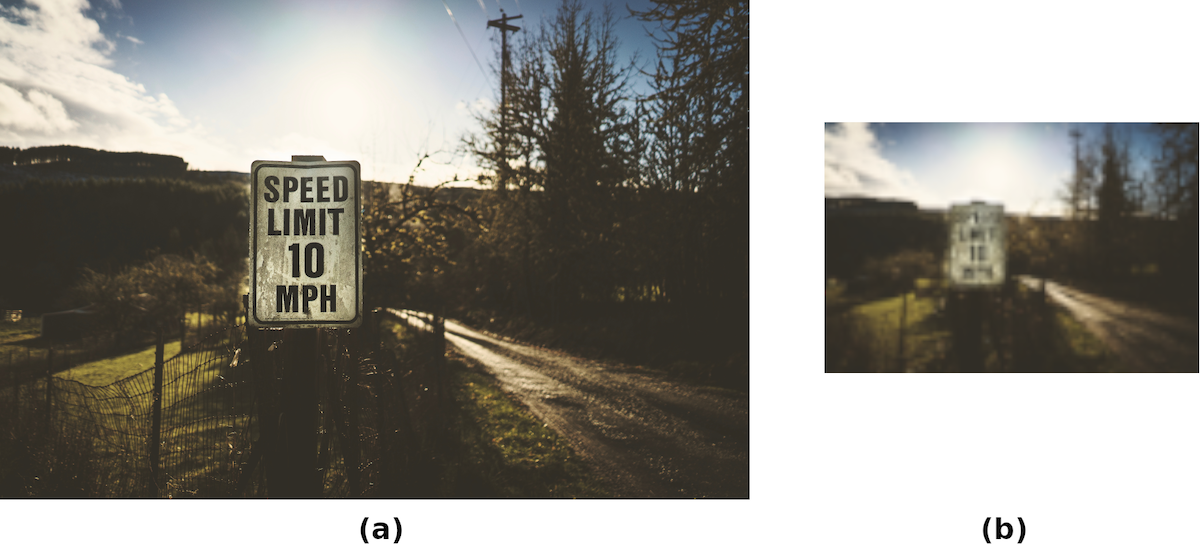}
    \caption{\textbf{(a)} Original large resolution
      image. \textbf{(b)} Downsampled image processable (in a
      reasonable time-frame) by a typical
      Resnet model using existing computational resources.}
    \label{problem}
    \end{center}
  \end{figure}

Direct inference over large input spaces allows models to leverage
fine grained information that might not be present in their downsampled
counterparts. We demonstrate a simple example of such a scenario in
Figure \ref{problem}, where the task is to identify speed
limits.
The downsampled image does not contain the required information to
correctly solve the task; on the other hand direct inference over the
original input space is memory and computationally intensive.



In order to work over such large dimensional input spaces, we take
inspiration from the way the human visual cortex handles high
dimensional input. Research in neuroscience \cite{rensink2000dynamic,
hayhoe2005eye, treisman1980feature} and attention for eye-gaze
\cite{ungerleider2000mechanisms} have suggested that human beings
enact rapid eye movements (or saccades \cite{deubel1996saccade}) to
different locations within the scene to gather high resolution
information from local patches.  More recent research
\cite{helfrich2018neural,fiebelkorn2018dynamic} has shown that humans
and macaque monkeys stochastically sample saccades from their
environment and merge them into a continuous representation of
perception. These saccades are also not necessarily only of the
salient object(s) in the environment, but have a component of
randomness attached to them. In this work we try to parallel this
stochastic element through the use of a learned sampling distribution,
conditioned on auxiliary information provided via a proxy
distribution. We explore two different types of proxy distributions in this work:
(a) one where the proxy distribution is simply the downsampled version of the original,
as in Figure \ref{problem} \& (b) one where the proxy distribution
is of a completely different modality. For the latter, we experiment
with a Starcraft II scenario \cite{vinyals2017starcraft}, and use the
game minimap as the proxy distribution.

\section{Related Work}

\textbf{Saliency Methods}: 
The analysis of salient (or interesting) regions in images has been
studied extensively in computer vision \cite{itti1998model,
  itti2000saliency, hou2007saliency, goferman2012context}. Important regions are quantified by simple
low-level features such as intensity, color and orientation
changes. These methods fail to generalize to complex scenes with
non-linear relationships between textures and colors \cite{bylinskii2016should}.  More recently,
deep convolutional networks have been exploited to directly learn
saliency at multiple feature levels (eg: \cite{li2015visual,
cao2018efficient}) as well as to learn patch level statistics
\cite{wang2015deep}. None of these methods directly learn ``where'' to look without information about the entire image.\\

\noindent\textbf{CNN Approaches}: Current state of the
art CNN models on the other hand separate the \emph{entire} image,
into cropped regions \cite{wu2017new}, employ pyramid decompositions
\cite{maggiori2017high} over the \emph{entire} image, or utilize large
pooling \cite{boureau2010theoretical} / striding operands.  These
methods are challenging because they are either lossy, resulting in
poor classification accuracy, or they are too memory and
computationally intensive (see Experiments Section \ref{experiments})
as they run convolutional filters over the entire image. \\

\noindent\textbf{Region Proposal Methods}: Another approach to CNN models are region proposal networks such as R-CNN
\cite{girshick2014rich}, R-CNN++ \cite{ren2015faster} and YOLO
\cite{redmon2016you} to name a few. The R-CNN methods generate a set of candidate
extraction regions, either by extracting a fixed number of proposals as in the
original work \cite{girshick2014rich}, or by utilizing a CNN over the \emph{entire image} to directly
predict the ROI \cite{ren2015faster}. They then proceed to enact a form
of pooling over these regions, compute features, and project the
features to the space of the classification likelihood. In contrast to
R-CNN, our method uses an informatively learned posterior to extract
the exact number of required proposals, rather than the 2000 proposals
as suggested in the original work. R-CNN++ on the other hand doesn't
scale with ultra-large dimensional images as direct inference over
these images scales with the dimensionality of the
images. Furthermore, the memory usage of R-CNN++ increases with the
dimensionality of the images whereas it does not for our proposed model.
YOLO on the other hand, resizes input images to $\bm{R}^{488 \times
488}$ and simultaneously predicts bounding boxes and their associated
probabilities. While YOLO produces quick classification
results, it trades off accuracy of fine-grained details. By resizing
the original image, critical information can be lost (see Figure \ref{problem}).
Our proposed method on the other hand has no trouble with small
details since it has the ability to directly control its foveation to
sample the full resolution image. \\

\noindent\textbf{Sequential Attention}: Sequential attention models have been
extensively explored through the literature, from utilizing Boltzman
Machines \cite{larochelle2010learning, denil2012learning,
  bazzani2011learning}, enacting step-by-step CNN learning
rules \cite{ranzato2014learning}, to learning
scanning policies \cite{alexe2012searching,butko2009optimal} as well
as leveraging regression based targets \cite{held2016learning}. Our
model takes inspiration from the recent Attend-Infer-Repeat (AIR) \cite{eslami2016attend} and its extensions (SQAIR) \cite{kosiorek2017hierarchical},
D.R.A.W \cite{gregor2015draw}, and Recurrent Attention Models (RAM) \cite{mnih2014recurrent,
  ba2014multiple}. While RAM based models allow for inference over
large input images, they utilize a score function
estimator \cite{glynn1990likelihood} coupled
with control variates \cite{glasserman2013monte}. Our algorithm on the
other hand utilizes pathwise estimators \cite{williams1992simple, kingma2014}
which have been shown to have lower variance
\cite{tokui2017evaluating} in practice. In contrast to AIR and general attention based
solutions, we do not use the entire image to build our attention
map. Our model can infer where to attend using a summary with
different semantics or encoding than the original image distribution (see Experiment \ref{sc2_exp}).
In addition, as opposed to adding a classifier in an ad-hoc manner as
in AIR and SQAIR, we derive a new principled lower bound on the
conditional classification likelihood that allows us to relate the
posterior of the proxy-distribution to the co-ordinate space of the
original input. This direct use of supervised information in an
end-to-end manner allows our model to converge very rapidly (100-300
epochs) vs AIR which takes 50,000-200,000 epochs
\cite{eslami2016attend} to successfully converge. \\

\noindent\textbf{Interpretability}: With the surge of deep learning, understanding
the model decision making process has become more important.
While prior work took a post-mortem approach on trained models
by computing gradients of the conditional likelihood with respect to
the input image
\cite{zeiler2014visualizing,mahendran2015understanding,
simonyan2013deep, olah2017feature}, recent work such as Capsule
Networks \cite{sabour2017dynamic}, InfoGAN \cite{chen2016infogan}, and
numerous others
\cite{sabour2017dynamic,zhang2017growing,wu2017interpretable,chen2016infogan}
directly attempt to learn models that are interpretable\footnote{See
\cite{zhang2018visual} for a more thorough treatment of
interpretability.}. Our model attempts to follow the latter of the two
paradigms by extracting crops of regions that directly maximize the
conditional classification likelihood. In contrast to the existing
methods mentioned above we do not parse the entire input image to provide
interpretability.


\section{Variational Objective} \label{variational_objective}

\begin{figure}[H]
  \begin{center}

    \includegraphics[width=\linewidth]{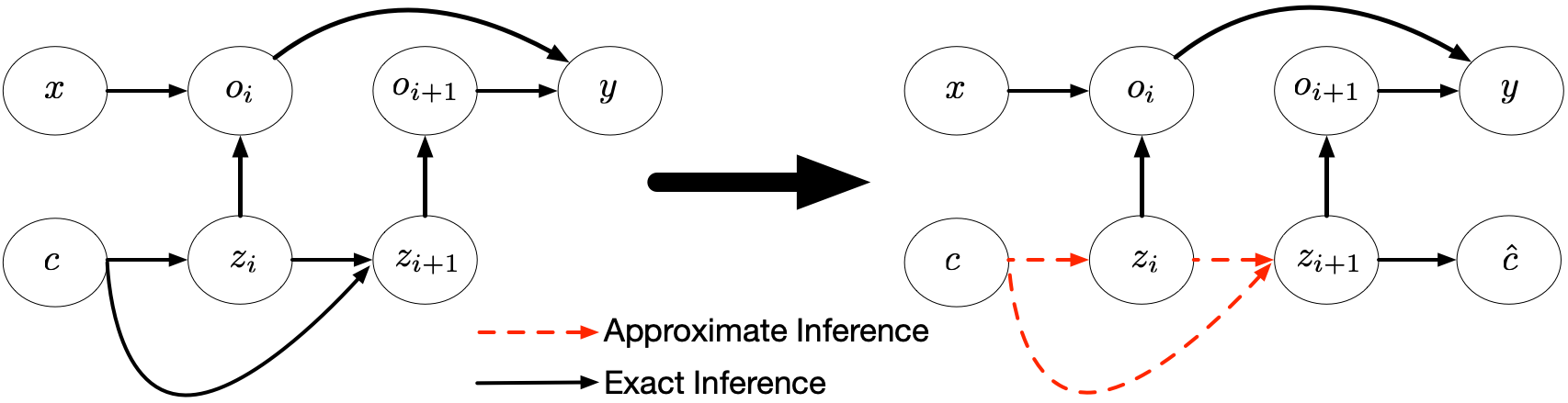}

    \caption{\emph{Left}: Original (intractable)
    objective (Equation \ref{eqn1}). \emph{Right}: Newly derived tractable objective
    (Equation \ref{eqn4}) that leverages a VRNN to approximate
    posteriors (red-dashed lines).}
    \label{graphical_model}
    \end{center}
  \end{figure}
\vspace{-0.2in}
Given an image $\bm{x} \in \mathbb{R}^{K \times K}$, a corresponding
proxy image $\bm{c} \in \mathbb{R}^{J \times J}$, $J \ll K$, and a
corresponding class label $\bm{y} \in \mathbb{R}$, our objective is
defined as maximizing $\log p_{\bm{\theta}}(\bm{y} | \bm{x})$ for
$\bm{\theta}$. 
We are only interested in the
case where $p(\bm{y}|\bm{c}) \neq p(\bm{y}|\bm{x})$, i.e. the proxy
distribution is not able to solve the classification task of
interest 
. Assuming that $\bm{c}$ provides no new
information for the classification objective, 
$p_{\bm{\theta}}(\bm{y}\ |\ \bm{x}) = p_{\bm{\theta}}(\bm{y}\ |\ \bm{x}, \bm{c})$,
we can
reformulate our objective as:

\begin{align}
\log p_{\bm{\theta}}(\bm{y} | \bm{x}) = \log p_{\bm{\theta}}(\bm{y} | \bm{x}, \bm{c}) =  \log \int \int
                                                  \frac{p_{\bm{\theta}}(\bm{y}
                                                  , \bm{o}_{\leq T},
                                                  \bm{z}_{\leq T},
                                                  \bm{c},
                                                  \bm{x})}{p(\bm{c},
                                                  \bm{x})}
                                                  d \bm{z}_{\leq T}
                                                  d \bm{o}_{\leq T}\label{eqn1}
\end{align}

We have introduced (and marginalized out) two sets of $T$ latent variables
in Equation \eqref{eqn1}: $\{\bm{z}\}_{i=1}^T, \bm{z}_i \in
\mathbb{R}^{3}$ and $\{\bm{o}\}_{i=1}^T, \bm{o}_i \in
\mathbb{R}^{L \times L}$, $L \ll J$. These correspond to the posteriors $\bm{z}_i \sim
p_{\bm{\phi}}(\bm{z}_i | \bm{c}, \bm{z}_{<i})$, induced by $\bm{c}$ and a set of dirac distributions, $\bm{o}_i
\sim \delta[f_{ST}(\bm{x}, \bm{z}_i)]$, centered at a differentiable
function, $f_{ST}$, implemented using Spatial Transformer
networks (ST) \cite{jaderberg2015spatial}\footnote{See Appendix
  Sections \ref{traditional_st} and \ref{local_spatial_xformer} for more
  information about Spatial Transformers and a possible method for
  minimizing their memory usage.}. This differentiable function
produces crops, $\bm{o}_i$, of our large original input, $\bm{x}$,
using a posterior sample from $p_{\bm{\phi}}(\bm{z}_i | \bm{c},
\bm{z}_{<i})$.
In general the true posterior, $p_{\bm{\phi}}(\bm{z}_i | \bm{c}, \bm{z}_{<i})$, is
intractable or difficult to approximate \cite{Kingma_undated-gm}. To resolve
this we posit a set of variational approximations \cite{wainwright2008graphical}, $q_{\bm{\phi}}(\bm{z}_i | \bm{c}, \bm{z}_{<i}) \approx p_{\bm{\phi}}(\bm{z}_i | \bm{c}, \bm{z}_{<i})$, and introduce them via
a multiply-by-one constant into the expanded joint distribution
implied by the graphical model in Figure \ref{graphical_model} \footnote{Full derivation in Appendix Section \ref{derivation}.}:

\begin{align}
  \begin{split}
  \log p_{\bm{\theta}}(\bm{y} | \bm{x}) = \log \int \int p_{\bm{\theta}_y}(\bm{y}|  \bm{o}_{\leq T}) \
    p_{\bm{\theta}_o}(\bm{o}_{\leq T} | \bm{z}_{\leq T}, \bm{x})
                                                  \
    p_{\bm{\phi}}(\bm{z}_{\leq T} | \bm{z}_{< T}, \bm{c})\  \frac{q_{\bm{\phi}}(\bm{z}_{\leq T} | \bm{z}_{< T},
    \bm{c})}{ q_{\bm{\phi}}(\bm{z}_{\leq T} | \bm{z}_{< T},
    \bm{c})} d \bm{z}_{\leq T} d \bm{o}_{\leq T}
\end{split}\label{eqn2}
\end{align}

By applying  Jensen's inequality and re-framing the marginalization
operand as an expectation we can rewrite Equation \ref{eqn2} from
above as a lower bound of $\log p_{\bm{\theta}}(\bm{y} | \bm{x})$:

\begin{align}
  \begin{split}
  \log p_{\bm{\theta}}(\bm{y} | \bm{x}) \geq \int \bigg[\mathbb{E}_{q_{\bm{\phi}}} \bigg(\log \bigg[
p_{\bm{\theta}_y}(\bm{y} | \bm{o}_{\leq T})\
                                                                  p_{\bm{\theta}_o}(\bm{o}_{\leq
                                                                  T} |
                                                                  \bm{z}_{\leq
                                                                  T},
                                                                  \bm{x})
                                                                  \bigg]\bigg)
                                                                  \\-
                                                                  \mathrm{D}_{KL}[q_{\bm{\phi}}(\bm{z}_{\leq
                                                                  T} |
                                                                  \bm{z}_{<
                                                                  T},
                                                                  \bm{c})
                                                                  ||
                                                                  p_{\bm{\phi}}(\bm{z}_{\leq
                                                                  T} |
                                                                  \bm{z}_{<
                                                                  T},
                                                                \bm{c})] \bigg]d \bm{o}_{\leq T}
                                                                \end{split}\label{eqn3}
\end{align}

We also observe that the KL divergence between the true set of
posteriors $p_{\bm{\phi}}(\bm{z}_{\leq T} | \bm{z}_{<T}, \bm{c})$ and the approximate
posteriors $q_{\bm{\phi}}(\bm{z}_{\leq T} | \bm{z}_{<T}, \bm{c})$ can be re-written in
terms of the VRNN \cite{chung2015recurrent} Evidence Lower
BOund (ELBO) \cite{kingma2014, chung2015recurrent} and the marginal data distribution
$p(\bm{c})$:

\begin{align}
      \begin{split}
 - D_{KL}[q_{\bm{\phi}}(&\bm{z}_{\leq T} | \bm{c}, \bm{z}_{<T})
       || p_{\bm{\phi}}(\bm{z}_{\leq T} | \bm{c}, \bm{z}_{<T})] \\&\
       \ \ \ \ \ \ \ \ \ \ = \\[-0.9ex]\mathbb{E}_{q_{\bm{\phi}}} \bigg(
  \sum_{i = 1}^T \log p_{\bm{\theta}}(\hat{\bm{c}} | \bm{z}_{\leq i})
  &- D_{KL}(q_{\bm{\phi}}(\bm{z}_i | \bm{c}, \bm{z}_{<i})
       || p_{\bm{\theta}}(\bm{z}_i | \bm{c}, \bm{z}_{<i})))
       \bigg) - \log p(\bm{c})
     \end{split}\label{ELBO_unrolled}
\end{align}

$p_{\bm{\theta}}(\bm{z}_i | \bm{c}, \bm{z}_{<i})$ in Equation
\ref{ELBO_unrolled} refers to the learned prior introduced by the VRNN,
while $\hat{\bm{c}} \sim \log p_{\bm{\theta}}(\hat{\bm{c}} |
\bm{z}_{\leq i})$ is the VRNN reconstruction. All temporal
dependencies $<T$ are incorporated by passing the hidden state of an
RNN across functions (see \cite{chung2015recurrent} for a more
thorough treatment). Coupling
the VRNN ELBO with the fact that $- \log p(\bm{c})$ is always a
positive constant, we can preserve the bound from Equation \ref{eqn3}
and update our reframed objective as:
\begin{align}
  \begin{split}
    \log p_{\bm{\theta}}(\bm{y} | \bm{x}) \gtrapprox \bigg[\mathbb{E}_{q_{\bm{\phi}}}\bigg( \log p_{\bm{\theta}_y}(\bm{y} |
                                                                                f_{\bm{\theta}_{\text{conv}}}(\underbrace{f_{ST}(\bm{z}_1,\bm{h}_1,
                                                                                \bm{x})}_{\bm{o}_1},...,
                                                                                \underbrace{f_{ST}(\bm{z}_T,
                                                                                \bm{h}_T,
                                                                                \bm{x})}_{\bm{o}_T})
                                                                                )
                                                                                \bigg)
     \\+\hspace{1.8in}\\\underbrace{\mathbb{E}_{q_{\bm{\phi}}} \bigg(
  \sum_{i = 1}^T \log p_{\bm{\theta}}(\hat{\bm{c}} | \bm{z}_{\leq i})
  - D_{KL}(q_{\bm{\phi}}(\bm{z}_i | \bm{c}, \bm{z}_{<i})
       || p(\bm{z}_i | \bm{c}, \bm{z}_{<i})))
       \bigg)}_{\text{VRNN
                                                                                  ELBO}} \bigg]
                                                                            \end{split}\label{eqn4}
\end{align}

This leads us to our final optimization objective, shown above in Equation \eqref{eqn4},
which utilizes a empirical estimate of the expectation and marginalization
operands and the substitutions of the functional approximations of the
dirac distributions, $\bm{o}_i \sim \delta[f_{ST}(\bm{x}, \bm{z}_i)]$, to provide a novel lower bound on $\log p_{\bm{\theta}}(\bm{y} |
\bm{x})$. This lower bound allows us to classify a set of crops, $\{\bm{o}_i\}_{i=1}^T$, of the
original distribution, $p(\bm{x})$, utilizing location information
inferred by the posterior, $q_{\bm{\phi}}(\bm{z}_{\leq T} | \bm{c}, \bm{z}_{<T})$, of the
proxy distribution, $p(\bm{c})$.

\subsection{Interpretation}
Current state of the art research in neuroscience for
attention \cite{helfrich2018neural,fiebelkorn2018dynamic} suggest
that humans sample saccades approximately every 250ms and integrate them into a
continuous representation of perception. We parallel this within our
model by utilizing a discrete $\bm{o}_i \sim \delta[f(\bm{x},
\bm{z}_i)]$ for sampling saccades and continuous latent
representations (isotropic-gaussian posteriors) for the
concept of perception. An additional requirement is the ability to
transfer this continuous latent representation across saccades. This
is enabled through the use of the VRNN, which relays information
between posteriors through its RNN hidden state.

In addition, \cite{helfrich2018neural,fiebelkorn2018dynamic} show that attention
does not always focus on the most salient object in an image, but at
times randomly attends to other parts of the scene. This behavior can be
interpreted as a form of exploration as done in reinforcement learning. In
our work, since the sampling distribution $q_{\bm{\phi}}(\bm{z}_i |
\bm{c}, \bm{z}_{<i})$ is stochastic, it provides a natural way to explore
the space of the original input distribution $p(\bm{x})$, without the
need for specific exploration methods such as $\epsilon$-greedy \cite{sutton1998introduction} or
weight noise \cite{fortunato2018noisy}. We validate this in Experiment \ref{ablation_studies}.


\section{Experiments}\label{experiments}
\begin{figure}[H]
  \includegraphics[width=\linewidth]{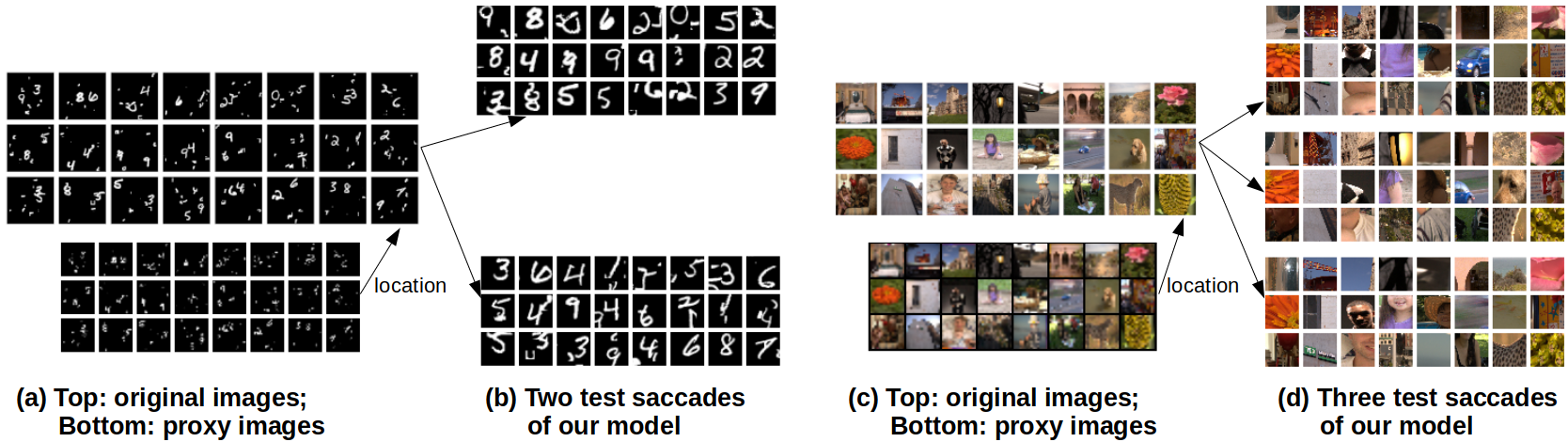}
  \caption{\emph{(a,b)}: Two-Digit-Identification ClutteredMNIST $\in
    \mathbb{R}^{2528 \times 2528}$; \emph{(c,d)}: MIT-5k $\in
    \mathbb{R}^{2528 \times 2528}$. Listed are test and proxy images (\emph{a,c})
    and their extracted test saccades (\emph{b,d}).}\label{results}
  \end{figure}

\begin{center}
\begin{table}[H]
  \scalebox{0.85}{%

\bgroup
\def\arraystretch{1.12}%

\begin{tabular}{llllll}
\cline{2-6}
\multicolumn{1}{l|}{\textbf{Image Size: $\mathbb{R}^{2528 \times 2528}$}}  & \multicolumn{1}{l|}{\textbf{\#params}} & \multicolumn{1}{l|}{\textbf{\begin{tabular}[c]{@{}l@{}}gpu memory\\ (160 batch-size)\end{tabular}}} & \multicolumn{1}{l|}{\textbf{time / epoch}}                                                             & \multicolumn{1}{l|}{\textbf{\begin{tabular}[c]{@{}l@{}}accuracy \\ MIT-Adobe-5k\end{tabular}}}       & \multicolumn{1}{l|}{\textbf{\begin{tabular}[c]{@{}l@{}}accuracy Two Digit\\ MNIST Identification\end{tabular}}} \\ \hline
\multicolumn{1}{|l|}{\textbf{resnet18 - 144 crops}}  &
                                                       \multicolumn{1}{l|}{11.4M}    & \multicolumn{1}{l|}{\begin{tabular}[c]{@{}l@{}}79G  (naive)\\ 6.5G (checkpoint)\protect\footnotemark \end{tabular}}       & \multicolumn{1}{l|}{\begin{tabular}[c]{@{}l@{}}1052.43s (naive) \\ 1454.05s (checkpoint)\end{tabular}} & \multicolumn{1}{l|}{\textbf{63.6\% +/- 0.03}}                                                        & \multicolumn{1}{l|}{\textbf{97.3 +/- 0.006}}                                                                    \\ \hline
\multicolumn{1}{|l|}{\textbf{variational saccading}} & \multicolumn{1}{l|}{\textbf{7.4M}}               & \multicolumn{1}{l|}{\textbf{4.1G}}                                                                    & \multicolumn{1}{l|}{\textbf{120 s}}                                                                    & \multicolumn{1}{l|}{62.7\% +/- 0.03}                                                                 & \multicolumn{1}{l|}{95.23 +/- 0.03}                                                                             \\ \hline
                                                     &                                        &                                                                                                     &                                                                                                        &                                                                                                      &                                                                                                                 \\ \cline{2-6}
\multicolumn{1}{l|}{\textbf{Image Size: $\mathbb{R}^{100 \times 100}$}}    & \multicolumn{1}{l|}{\textbf{\#params}} & \multicolumn{1}{l|}{\textbf{\begin{tabular}[c]{@{}l@{}}gpu memory\\ (160 batch-size)\end{tabular}}}                                                            & \multicolumn{1}{l|}{\textbf{time / epoch}}                                                             & \multicolumn{1}{l|}{\textbf{\begin{tabular}[c]{@{}l@{}}accuracy Two Digit\\ MNIST Sum\end{tabular}}} & \multicolumn{1}{l|}{\textbf{\begin{tabular}[c]{@{}l@{}}accuracy Two Digit\\ MNIST Identification\end{tabular}}} \\ \hline
\multicolumn{1}{|l|}{\textbf{resnet18 - full image}} & \multicolumn{1}{l|}{11M}               & \multicolumn{1}{l|}{6.6G}                                                                          & \multicolumn{1}{l|}{59.27s}                                                                            & \multicolumn{1}{l|}{\textbf{99.86 +/- 0.01}}                                                         & \multicolumn{1}{l|}{\textbf{97.4 +/- 0.003}}                                                                    \\ \hline
\multicolumn{1}{|l|}{\textbf{RAM} \cite{mnih2014recurrent}}                   & \multicolumn{1}{l|}{\textbf{-}}                 & \multicolumn{1}{l|}{\textbf{-}}                                                                              & \multicolumn{1}{l|}{\textbf{-}}                                                                        & \multicolumn{1}{l|}{91\%}                                                                            & \multicolumn{1}{l|}{93\%}                                                                                       \\ \hline
\multicolumn{1}{|l|}{\textbf{DRAM} \cite{ba2014multiple}}                  & \multicolumn{1}{l|}{\textbf{-}}        & \multicolumn{1}{l|}{\textbf{-}}                                                                     & \multicolumn{1}{l|}{\textbf{-}}                                                                        & \multicolumn{1}{l|}{97.5\%}                                                                          & \multicolumn{1}{l|}{95\%}                                                                                       \\ \hline
\multicolumn{1}{|l|}{\textbf{variational saccading}} & \multicolumn{1}{l|}{\textbf{7.4M}}               & \multicolumn{1}{l|}{\textbf{2.8G}}                                                                           & \multicolumn{1}{l|}{\textbf{37s}}                                                                      & \multicolumn{1}{l|}{97.2 +/- 0.04}                                                                   & \multicolumn{1}{l|}{95.42 +/- 0.002}                                                                            \\ \hline
\end{tabular}

\egroup
}%
\vskip 0.1in
\caption{\small Our model infers $\sim \textbf{9-10x}$ faster and utilizes less GPU
  memory than the baselines in high dimensions. \emph{Top}: Large
  resolution trials on MIT-Adobe-5k and Two-Digit MNIST Identitification. \emph{Bottom}: Small resolution baseline trials
  used to situate work against RAM \cite{mnih2014recurrent} and DRAM\cite{ba2014multiple}.}\label{table_results}
\end{table}
\end{center}
\vspace{-0.4in}




\footnotetext{Checkpointing caches the forward pass operation as described
  in \cite{chen2016training}. The naive approach parallelizes across 8
  GPUs and splits each of the 144 crops across the GPUs.}

We evaluate our algorithm on three classification datasets where we
analyze different induced behaviors of our model. We utilize Two-Digit
MNIST for our first experiment in order to situate our model against
baselines; we then proceed to learn a classification model for the
large MIT-Adobe 5k dataset and finally attempt to learn a model that can
accuractely count marines in a complicated, dynamic, large resolution
Starcraft II \cite{vinyals2017starcraft} map. The first two experiments utilize downsampled
original images, $\bm{c} \in \mathbb{R}^{32 \times 32}$, as
the proxy distribution, while the third uses the game-minimap from
Starcraft II, $\bm{c} \in \mathbb{R}^{64 \times 64}$, as an auxiliary source of information.
 We demonstrate that our model has comparable accuracy to
the best baseline models in the first two experiments, but we infer $\sim \textbf{9-10x}$ faster and utilize far less
GPU memory than a naive approach. We provide visualizations of the
model's saccades in Figure \ref{results}; this aids in interpreting what region of the
original input image aids the model in maximizing the desired classification
likelihood. We utilize resnet18 as our naive baseline and did not
observe any performance uplifts from using larger models for our three experiments.

We implement the VRNN using a
fully convolutional architecture where conv-transpose layers are used
for upsampling from the vectorized latent space. The crop classifier is implemented by a standard fully-convolutional
network, followed by a spatial pooling operation on the results of the
convolution over the crops, $\bm{o}_i$. Adam \cite{kingma2014adam} was used as an optimizer, combined with ReLU
activations; batch-norm \cite{ioffe2015batch} was used for dense
layers and group-norm \cite{wu2018group} for convolutional
layers. For more details about specific architectural choices see
our code\footnote{\url{https://github.com/jramapuram/variational_saccading}} and Appendix Section \ref{model_desc} in
the supplementary material.


\subsection{Two-Digit MNIST: Identification \& Summing}\label{two_digit_mnist}

Two-Digit-Cluttered MNIST is a benchmark dataset used in RAM
\cite{mnih2014recurrent}, DRAM \cite{ba2014multiple} and as a
generative target in AIR \cite{eslami2016attend} and SQAIR
\cite{kosiorek2017hierarchical}\footnote{The authors do not use the
cluttered version of the two-digit dataset for the AIR variants.}. The
objective with the initial set of experiments is to identity the
digits present in the image (ignoring the distracting clutter), localize them, and
predict a multi-class label using the localized targets. This form of
learning, where localization information is not directly provided, is
known as weakly supervised learning \cite{carneiro2007supervised,
ramapuramnew,oquab2015object}. In the first setting we compare our
model to RAM \cite{mnih2014recurrent}, DRAM \cite{ba2014multiple} and
a baseline resnet18 \cite{he2016deep} model that operates over the
entire image and directly provides classification outputs. As in RAM
and DRAM, we also examine a case where the learning objective is
to sum two digits placed in an image (without clutter). In order to provide a fair comparison we evaluate our model in the original
dimension ($\mathbb{R}^{100 \times 100}$) suggested by the authors
\cite{mnih2014recurrent, ba2014multiple}. We observe (Table
\ref{table_results} bottom) that our method improves upon RAM and DRAM and gets close to the baseline resnet18 results.

We extend the Two-Digit-Cluttered MNIST identification experiment from
above to a new experiment where we classify large dimensional images, $\bm{x} \sim \mathbb{R}^{2528 \times
  2528}$. As in the previous experiment we evaluate our model against a baseline resnet18 model. Resnet models are tailored to operate over $\mathbb{R}^{224
  \times 224}$ images; in order to use large images, we divide an original $\mathbb{R}^{2528
  \times 2528}$ image into $\mathbb{R}^{144 \times 224
  \times 224}$ individual crops and feed each crop into the model. We
then sum the logit outputs of the model and run the pooled result
through a dense layer. This allows the model to make a single
classification decision for the entire image using all 144 crops: $\bm{y} = f_{\bm{\theta}_d}(\sum_{i=1}^{144}
  g_{\bm{\theta}_c}(\bm{x}_i))\ ,\ \bm{x}_i \in \mathbb{R}^{224 \times
  224}\label{resnet_alg}$.
$f_{\bm{\theta}_d}$ represents a multi-layer dense
network and $g_{\bm{\theta}_c}$ is a multi-layer convolutional neural
network that operates on individual crops $\bm{x}_i$. While it is
also possible to concatenate each logit vector
$g_{\bm{\theta}_c}(\bm{x}) = [g_{\bm{\theta}_c}(\bm{x}_i),\ g_{\bm{\theta}_c}(\bm{x}_{i-1}), ...,\ g_{\bm{\theta}_c}(\bm{x}_0)]$, and project it through the dense
network $f_{\bm{\theta}_d}(g_{\bm{\theta}_c}(\bm{x}))$, the tasks
we operate over do not necessitate relational information \cite{santoro2017simple} and pooled
results directly aid the classification objective. We visualize
saccades (Figure \ref{results}), the model accuracy,
training-time per epoch and GPU memory (Table \ref{table_results}) and
observe that our model performs similarly (in terms of accuracy) in
higher dimensions, while inferring \textbf{$\sim$10x}
faster and using \textbf{only 5\%} of the total GPU memory in contrast to a
traditional resnet18 model.

\subsubsection{Ablation Studies}\label{ablation_studies}

\begin{figure}[H]
  \begin{minipage}{0.49\textwidth}
    \includegraphics[width=\linewidth]{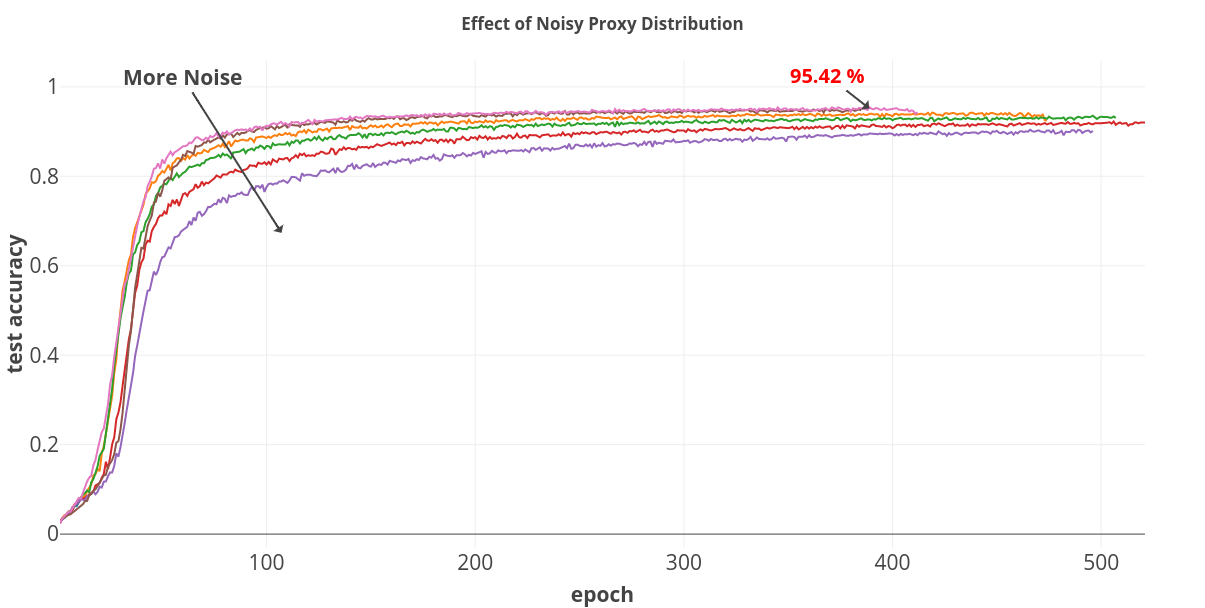}
    \includegraphics[width=\linewidth]{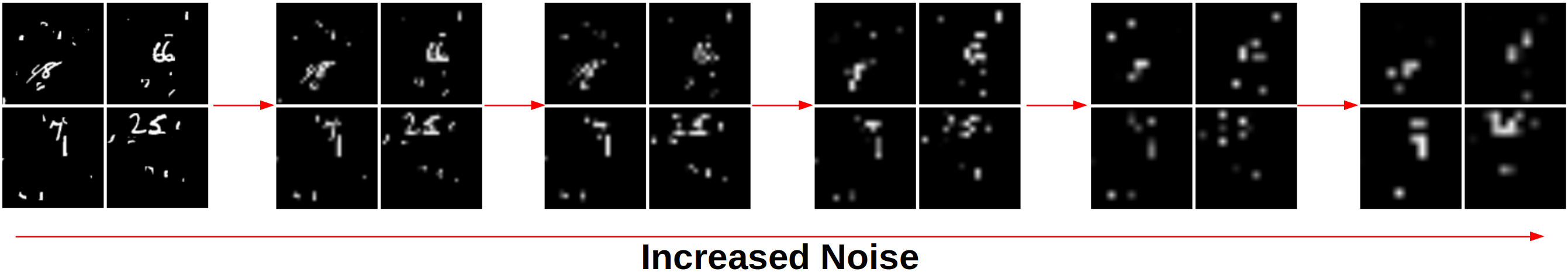}
\end{minipage}%
\begin{minipage}{0.49\textwidth}
    \includegraphics[width=\linewidth]{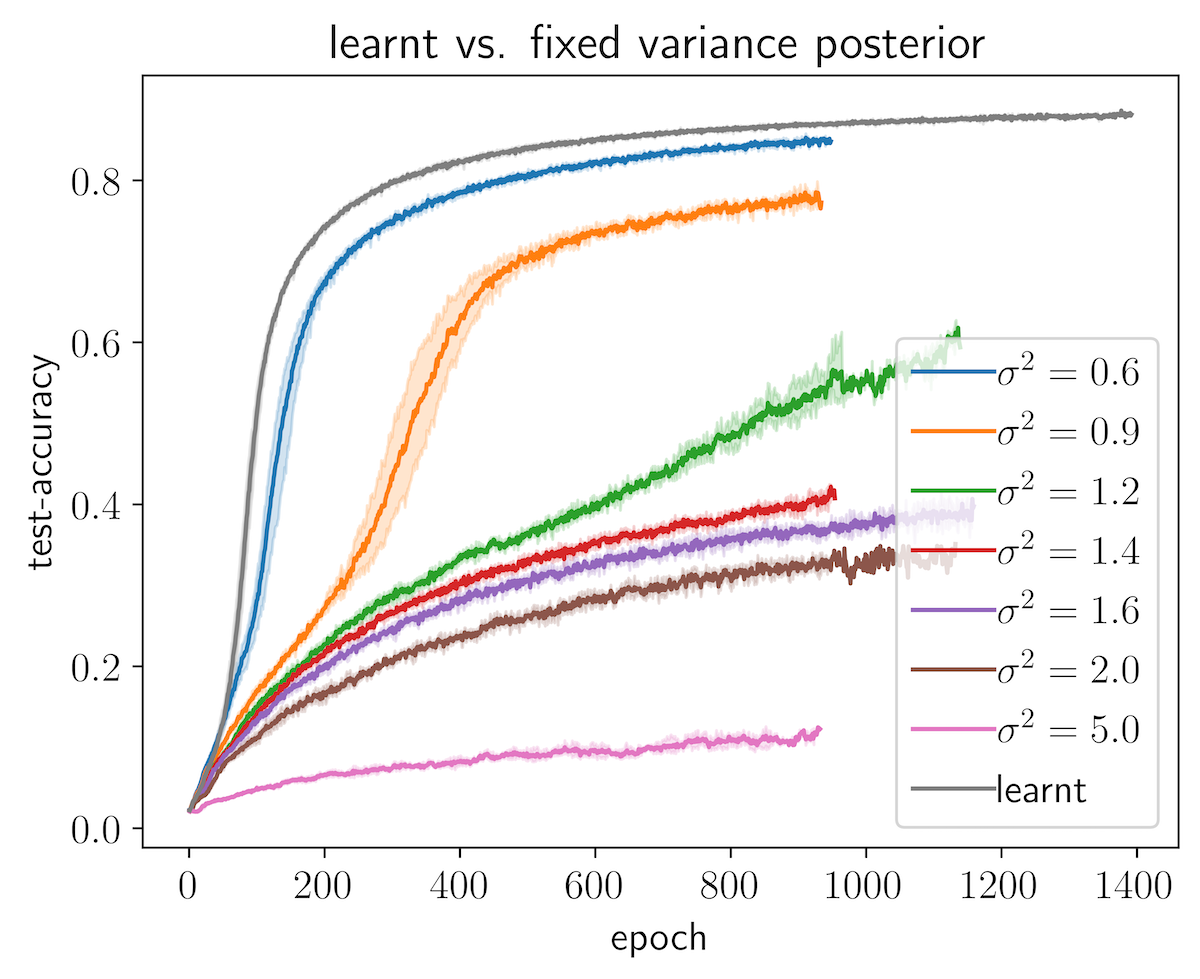}
\end{minipage}%
\caption{\emph{Top-left}: Effect of noisy proxy distribution on test
  accuracy. \emph{Bottom-left}: Left to right correspond to noisier
  versions of the same proxy distribution used in above
  graph. \emph{Right}: Test accuracy for Two-Digit ClutteredMNIST identification under a range of fixed variances,
    $\sigma^2$, of the isotropic-gaussian posterior $q_{\bm{\phi}}(\bm{z}_i | \bm{z}_{<i}, \bm{c})$.}\label{effect_downsampling}
\end{figure}

\noindent\textbf{Robustness to Noisy Proxy Distribution:} Since the proxy distribution is critical to our formulation, we
conduct an ablation study using the two-digit cluttered identification
problem from experiment \ref{two_digit_mnist}. We vary the amount of
noise in the proxy distribution as shown in the \emph{bottom} of Figure
\ref{effect_downsampling}-\emph{left}. The test curves shown on the \emph{top} of
the same figure demonstrates that our method is
robust to noisy proxy distributions. In general, we found
that our method worked even in situations where the proxy distribution
only contained a few points, allowing us to infer positional
information to index the original distribution $p(\bm{x})$.\\

\noindent \textbf{Quantifying Learned Exploration:} In order to validate the hypothesis that the learned variance, $\bm{\sigma}^2$,
of our isotropic-gaussian posterior $q_{\bm{\phi}}(\bm{z}_i |
\bm{z}_{<i}, \bm{c})$ is useful in the learning process, we repeat the identification
experiment from Experiment \ref{two_digit_mnist} using the noisiest
proxy-distribution from the previous ablation study (right most
example in Figure \ref{effect_downsampling}-\emph{bottom-left}). We compare our model
(\emph{learnt}) against the same model with varying fixed variance:
$\bm{\sigma}^2$ = [0.6I, 0.9I, 1.2I, 1.4I, 1.6I, 2.0I, 5.0I]. We
repeat each experiment five times and plot the mean and variance of
the test accuracy in Figure \ref{effect_downsampling}-\emph{right}. We observe a \textbf{clear} advantage in terms of
convergence time and accuracy for the \emph{learnt} model.

\subsection{MIT-Adobe 5k}
MIT-Adobe 5k \cite{fivek} is a high resolution DSLR camera dataset consisting
of six classes: \{abstract, animals, man-made, nature, None,
people\}. While the dimensionality of each image is large, the dataset
has a total of 5000 samples. This upper-bounds the performance of
deep models with millions of parameters (without the use of
pre-training / fine-tuning and other unsupervised techniques). We
examine this scenario because it presents a common use case of
learning in a low-sample regime.

We downsample the large original images to $\bm{x} \in \mathbb{R}^{3 \times 2528 \times 2528}$ to evaluate against a
baseline resnet \cite{he2016deep} model. The baseline model operates
over 144 crops per image
as in the previous experiment.
Test saccades (non-cherry picked) of our model are visualized in
Figure \ref{results}; the saccades allow us to gain an introspective
view into the model decision making process. Some of the interesting
examples are that of the `people' class: in the example with the child
(third to the right in the bottom row of Figure \ref{results}-c),
the model saccades to the adult as well as the child in the
image. Other notable examples are leveraging the spotted texture of
cheetah fur and the snout of the dog.
As observable from Table \ref{table_results}-\emph{top}, our model has comparable accuracy to the
baseline resnet model, but infers $\sim \textbf{9-10x}$ faster and
uses far less GPU memory than a naive approach.

\subsection{Starcraft II - Count the Marines}\label{sc2_exp}
\begin{figure}[H]
\minipage{0.4\linewidth}
  \includegraphics[width=\linewidth]{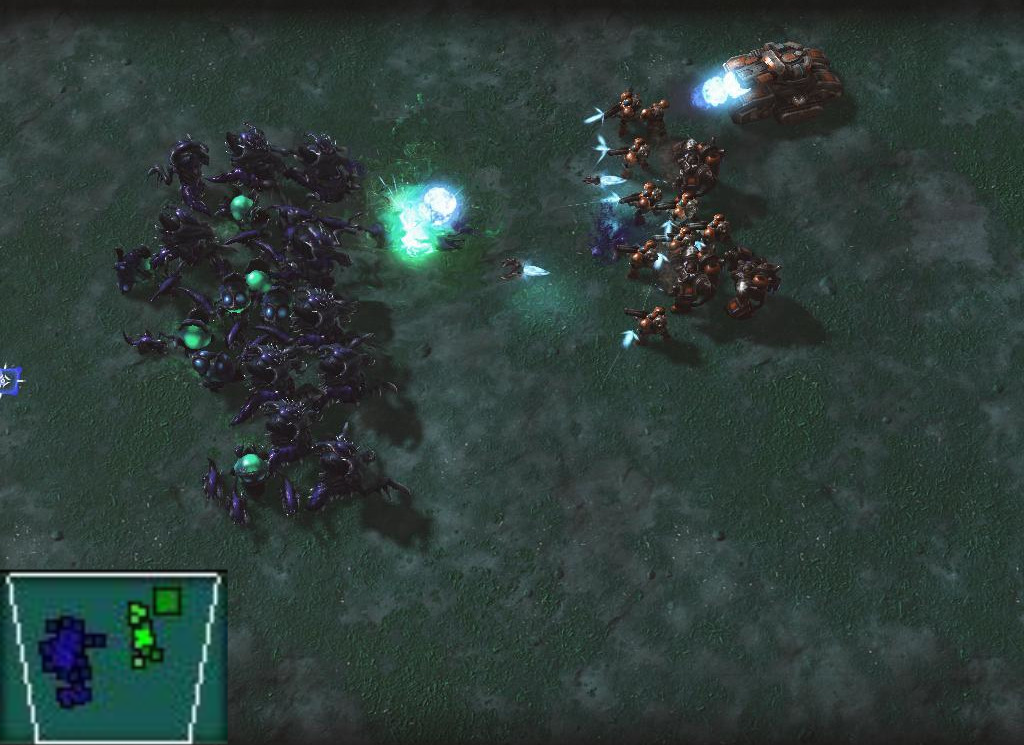}
\endminipage\hfill
\minipage{0.6\linewidth}
\includegraphics[width=\linewidth]{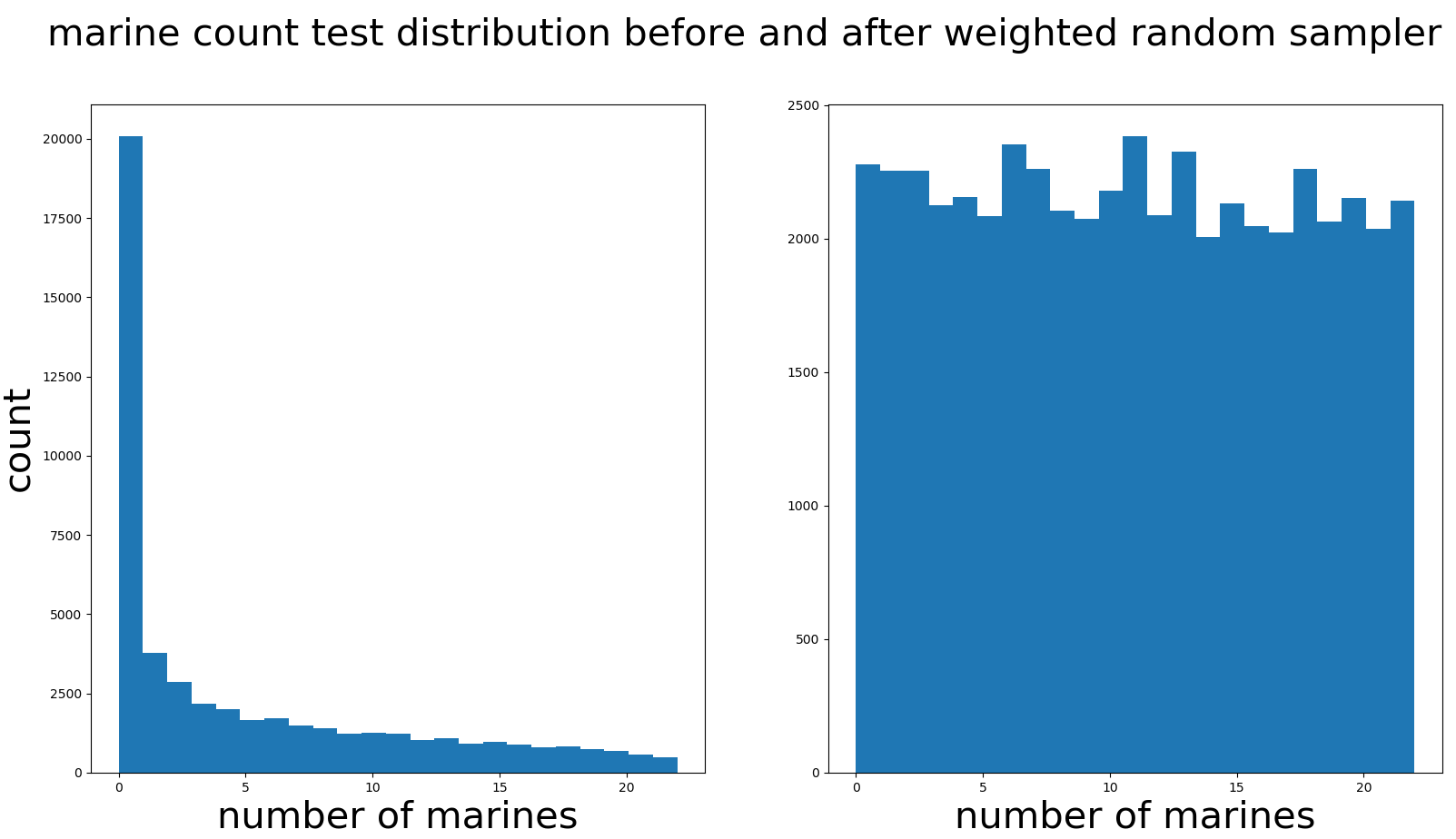}
  \endminipage\hfill
    \caption{\emph{Left}: full resolution image with minimap (proxy
      distribution, $p(\bm{c})$) in the bottom left corner. \emph{Right}:
    initial test output distribution and normalized test distribution using weighted random sampler with replacement.} \label{sc2}
\end{figure}

In this experiment we generate a dataset of 250,000 samples of a
randomly initialized Starcraft II \cite{vinyals2017starcraft}
scenario. The maps generated use the full resolution ($\bm{x} \in
\mathbb{R}^{1024 \times 1024}$) images which have complex structure
such as bloom, background texture and anti-aliased sprites (see Figure
\ref{sc2} \emph{left}).  Units from the Zerg faction attack Terran units until all
units of one faction are eliminated. Both factions generate a random
number of three distinct types of units with a maximum of 22 per unit-type. Our objective in this scenario
is to predict the number of Terran marines in
the map at any given time given training data with the correct counts;
we use the minimap (Figure \ref{sc2}-\emph{left} bottom corner) as our proxy
distribution, $p(\bm{c})$, to infer positional
information in the full-resolution image $p(\bm{x})$. Even though the minimap has
an estimate of the total count the model cannot directly use this
information to make a prediction due to the fact that \emph{there are multiple unit types with
  the same minimap grid size}. This forces the model to have to look
at the true map in order to infer the number of marines present. In
addition, the minimap is only used to infer locations in the
full-resolution images and not directly for classification.
\begin{figure*}
  \begin{center}
  \scalebox{0.8}{\includegraphics[width=\linewidth]{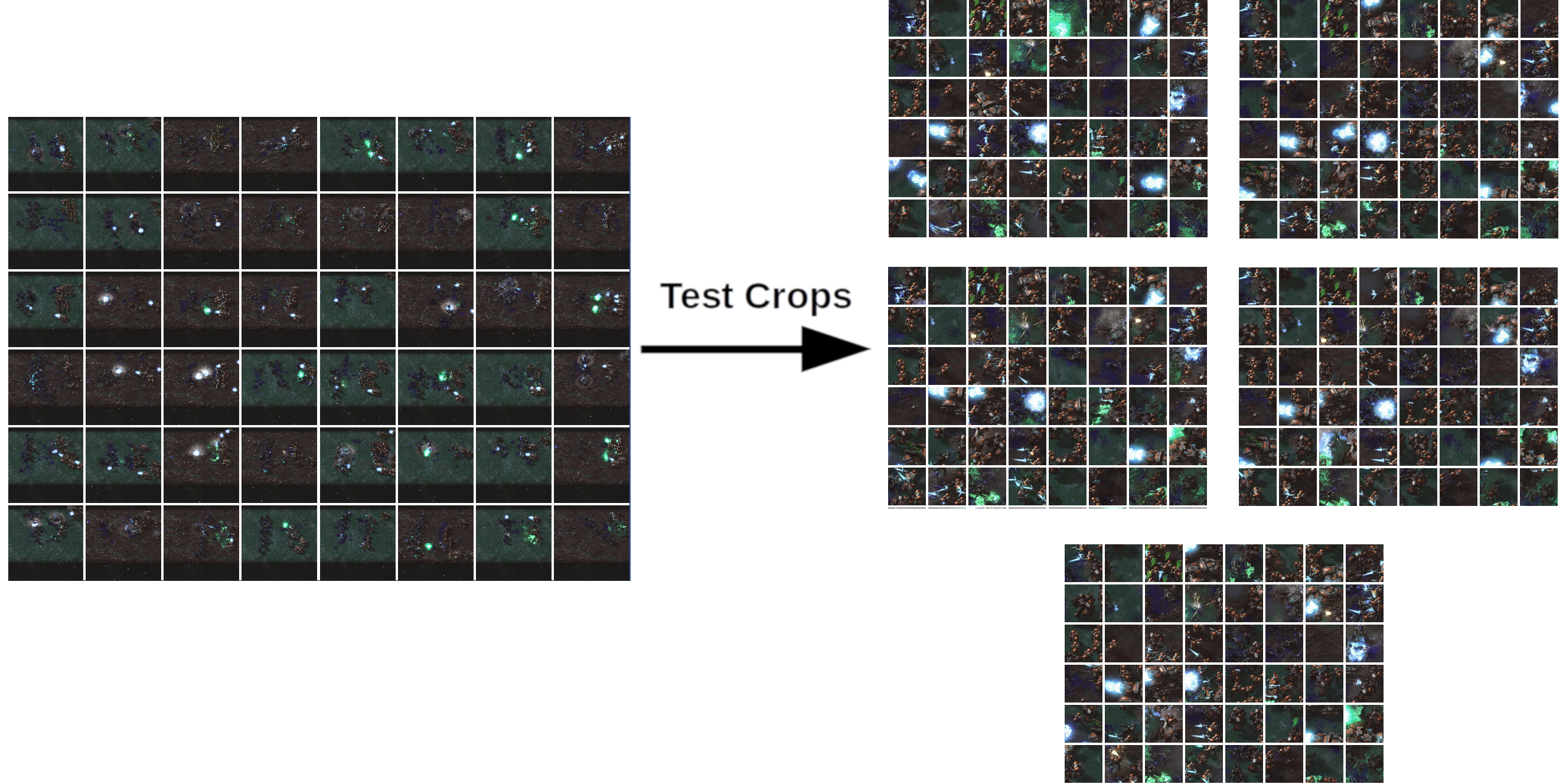}}
  \caption{\emph{Left}: Starcraft II full resolution test images $\in
    \mathbb{R}^{1024 \times 1024}$. \emph{Right}: Five Test saccades
    (per input image) extracted by our model; almost all contain localized Terran marines.}\label{sc2_results}
\end{center}
\end{figure*}

The generated dataset was initially biased to a count of zero marines;
in order to learn a model that did not instantly have $>$ 50\% accuracy we
used a weighted random sampler (with replacement) on both the train
and test sets. We visualize the effect of this sampler on the entire
test distribution in the histograms shown in Figure \ref{sc2}-\emph{right}.

As with the previous experiment we compare our model to a baseline
resnet18 that utilized the entire image to predict the Terran marine
count. In this experiment we observe that our model (50\%) does not perform
as well as the baseline resnet18 (65\%) in terms of classification accuracy,
however we observe reasonable localizations in Figure
\ref{sc2_results}. We hypothesize that this is due to the fact that we
use a completely different modality for the proxy distribution.
This makes it challenging for the model to
learn how to saccade around the space as it needs to first learn
the relationship of the simplistic color RGB mini-map and
the complicated, texture filled, full-resolution map. We hypothesize that
the reason for the performance degradation is due to 
double-counting of marines. 
Our work presents a strong first attempt at probabilistic large image
classification using location information derived from a proxy
image. The method also provides localization for free as a byproduct
of the process.



\section{Conclusion}
We demonstrate a novel algorithm capable of working with ultra-large
resolution images for classification and derive a new principled variational
lower bound that captures the relationship of a proxy distribution's posterior and the
original image's co-ordinate space
. We empirically demonstrate
that our model works with low
memory and inference costs on ultra-large images in three datasets.

\section{Acknowledgments}
We would like to thank Amazon Web Services (AWS) and the Swiss National
Supercomputing Centre (CSCS) for their generous GPU grants, without
which the experiments in this paper would not be possible. We would also like to thank
our hard-working systems administrator, Yann Sagon, from the University of
Geneva, who keeps our internal cluster operational.

{\small
\bibliographystyle{ieee}
\bibliography{variational_saccading}
}


\newpage
\section{Appendix}

\subsection{Lower Bound Derivation}\label{derivation}

\scalebox{1.0}{\parbox{1.0\linewidth}{%
    \begin{align}
      \begin{split}
\log p_{\bm{\theta}}(\bm{y} | \bm{x}) &= \log \int \int \frac{p_{\bm{\theta}_y}(\bm{y}|  \bm{o}_{\leq T}) \prod_{i=1}^T[
                                                  \ p_{\bm{\theta}_o}(\bm{o}_{i} | \bm{z}_{i}, \bm{x})
                                                  \
    p_{\bm{\phi}}(\bm{z}_i | \bm{z}_{< i}, \bm{c})\  d \bm{z}_{i} ]\ \cancel{p(\bm{c},
                                                  \bm{x})}}{\cancel{p(\bm{c},
                                                  \bm{x})}}
    \frac{\prod_{i=1}^T q_{\bm{\phi}}(\bm{z}_{i} | \bm{z}_{< i},
    \bm{c})}{\prod_{i=1}^T q_{\bm{\phi}}(\bm{z}_{i} | \bm{z}_{< i},
    \bm{c})} d \bm{o}_{\leq T} \\ &= \log \int \int p_{\bm{\theta}_y}(\bm{y}|  \bm{o}_{\leq T}) \
    p_{\bm{\theta}_o}(\bm{o}_{\leq T} | \bm{z}_{\leq T}, \bm{x})
                                                  \
    p_{\bm{\phi}}(\bm{z}_{\leq T} | \bm{z}_{< T}, \bm{c})\  \frac{q_{\bm{\phi}}(\bm{z}_{\leq T} | \bm{z}_{< T},
    \bm{c})}{ q_{\bm{\phi}}(\bm{z}_{\leq T} | \bm{z}_{< T},
    \bm{c})} d \bm{z}_{\leq T} d \bm{o}_{\leq T}\\
   &\geq \int \int q_{\bm{\phi}}(\bm{z}_{\leq T} | \bm{z}_{< T},
    \bm{c}) \log \bigg[p_{\bm{\theta}_y}(\bm{y}|  \bm{o}_{\leq T}) \
    p_{\bm{\theta}_o}(\bm{o}_{\leq T} | \bm{z}_{\leq T}, \bm{x})
                                                  \
    p_{\bm{\phi}}(\bm{z}_{\leq T} | \bm{z}_{< T}, \bm{c})\  \frac{1}{ q_{\bm{\phi}}(\bm{z}_{\leq T} | \bm{z}_{< T},
  \bm{c})} d \bm{z}_{\leq T} d \bm{o}_{\leq T}\bigg] \\
  &= \int \bigg( \mathbb{E}_{q_{\bm{\phi}}} \bigg[\log p_{\bm{\theta}_y}(\bm{y}|  \bm{o}_{\leq T}) \
    p_{\bm{\theta}_o}(\bm{o}_{\leq T} | \bm{z}_{\leq T}, \bm{x})\bigg]
    - \int q_{\bm{\phi}}(\bm{z}_{\leq T} | \bm{z}_{< T}, \bm{c}) \log
                                                  \
    \frac{ q_{\bm{\phi}}(\bm{z}_{\leq T} | \bm{z}_{< T},
      \bm{c})}{p_{\bm{\phi}}(\bm{z}_{\leq T} | \bm{z}_{< T}, \bm{c})}
    d \bm{z}_{\leq T} \bigg) d \bm{o}_{\leq T}\\
    &= \int \bigg( \mathbb{E}_{q_{\bm{\phi}}} \bigg[\log p_{\bm{\theta}_y}(\bm{y}|  \bm{o}_{\leq T}) \
    p_{\bm{\theta}_o}(\bm{o}_{\leq T} | \bm{z}_{\leq T}, \bm{x})\bigg]
    - D_{KL}[q_{\bm{\phi}}(\bm{z}_{\leq T} | \bm{c}, \bm{z}_{<T}) ||
    p_{\bm{\phi}}(\bm{z}_{\leq T} | \bm{c}, \bm{z}_{<T})] \bigg) d
    \bm{o}_{\leq T}\\
    &= \int \bigg( \mathbb{E}_{q_{\bm{\phi}}} \bigg[\log p_{\bm{\theta}_y}(\bm{y}|  \bm{o}_{\leq T}) \
    p_{\bm{\theta}_o}(\bm{o}_{\leq T} | \bm{z}_{\leq T}, \bm{x})\bigg]
    \\&\hspace{0.2in}+ \mathbb{E}_{q_{\bm{\phi}}} \bigg(
  \log p_{\bm{\theta}}(\hat{\bm{c}} | \bm{z}_{\leq T})
  - D_{KL}(q_{\bm{\phi}}(\bm{z}_{\leq T} | \bm{c}, \bm{z}_{< T})
       || \underbrace{p_{\bm{\theta}}(\bm{z}_{\leq T} | \bm{c},
         \bm{z}_{< T})}_{\text{VRNN prior}}))
       \bigg) - \log p(\bm{c}) \bigg) d
       \bm{o}_{\leq T}\\
       &\geq \int \bigg( \mathbb{E}_{q_{\bm{\phi}}} \bigg[\log p_{\bm{\theta}_y}(\bm{y}|  \bm{o}_{\leq T}) \
    p_{\bm{\theta}_o}(\bm{o}_{\leq T} | \bm{z}_{\leq T}, \bm{x})\bigg]
    \\&\hspace{0.2in}+ \mathbb{E}_{q_{\bm{\phi}}} \bigg(
  \log p_{\bm{\theta}}(\hat{\bm{c}} | \bm{z}_{\leq T})
  - D_{KL}(q_{\bm{\phi}}(\bm{z}_{\leq T} | \bm{c}, \bm{z}_{< T})
       || p_{\bm{\theta}}(\bm{z}_{\leq T} | \bm{c}, \bm{z}_{< T})))
       \bigg) \bigg) d
       \bm{o}_{\leq T} \\
       &\approx \mathbb{E}_{q_{\bm{\phi}}}\bigg( \log p_{\bm{\theta}_y}(\bm{y} |
                                                                                f_{\bm{\theta}_{\text{conv}}}(\underbrace{f_{ST}(\bm{z}_1,\bm{h}_1,
                                                                                \bm{x})}_{\bm{o}_1},...,
                                                                                \underbrace{f_{ST}(\bm{z}_T,
                                                                                \bm{h}_T,
                                                                                \bm{x})}_{\bm{o}_T})
                                                                                )
                                                                                \bigg)
     \\&\hspace{1.8in}+\\ &\hspace{5mm}\underbrace{\mathbb{E}_{q_{\bm{\phi}}} \bigg(
  \sum_{i = 1}^T \log p_{\bm{\theta}}(\hat{\bm{c}} | \bm{z}_{\leq i})
  - D_{KL}(q_{\bm{\phi}}(\bm{z}_i | \bm{c}, \bm{z}_{<i})
       || p(\bm{z}_i | \bm{c}, \bm{z}_{<i})))
       \bigg)}_{\text{VRNN
                                                                                  ELBO}}
    \end{split}\nonumber
\end{align}
}}



\subsection{Traditional Spatial Transformer}\label{traditional_st}
To produce the crops $\bm{o}_i$, we utilize Spatial Transfomers (ST)
\cite{jaderberg2015spatial}.
STs transform the
process of hard-attention based cropping (i.e. indexing into the
image) with two differentiable operators: a learned affine transformation
of the \emph{co-ordinate space} of the original image,
$\begin{bmatrix} i^s& j^s \end{bmatrix}^T
\mapsto \begin{bmatrix} i^t& j^t \end{bmatrix}^T$:

\begin{align}
\begin{bmatrix}i^t \\ j^t \end{bmatrix} = \begin{bmatrix} s
  & 0 & x \\ 0& s& y \end{bmatrix} \begin{bmatrix} i^s \\ j^s \\ 1 \end{bmatrix} = \begin{bmatrix} z_0 & 0 & z_1 \\
  0 & z_0 & z_2 \end{bmatrix} \begin{bmatrix} i^s \\ j^s \\ 1 \end{bmatrix}
\end{align}

and a differentiable bilinear sampling operator that is
independently applied on each channel $c$:
\begin{align}
  \sum_n^J \sum_m^J \bigg( x^c_{nm} \max (0, 1-|i^t_{nm} -
  m|) \nonumber \max (0, 1-|j^t_{nm} - n|) \bigg)
\end{align}

\subsection{Localized Spatial
  Transformer}\label{local_spatial_xformer}

\begin{figure}[H]
\begin{center}
  \centerline{\includegraphics[width=\linewidth]{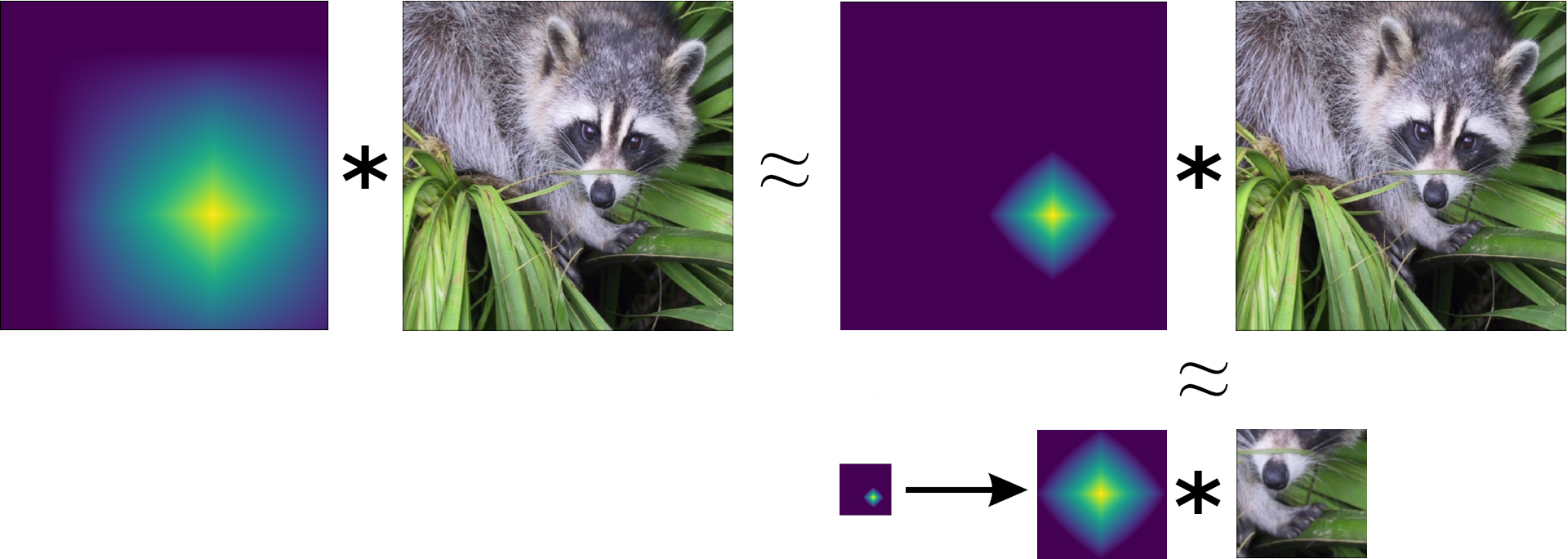}}
  \caption{Localized spatial transformers allow us to directly crop
    the ROI from an image without loading the entire image or
    generating the full flow-field. A flow field is generated and
    truncated in a low dimensional resolution, cropped to the ROI, and
    upsampled to the desired crop size.} \label{trunc_st}
  \vspace{-0.3in}
  \end{center}
\end{figure}

Spatial Transfomers (ST) \cite{jaderberg2015spatial}, transform the
process of hard-attention based cropping (i.e. indexing into the
image) with two differentiable operators: 1) a learned affine transformation
of the co-ordinate space of the original image and 2) a differentiable
bilinear sampling operator as shown in Equation
\ref{diff_sampling}. This sampling operator generally requires the
entire input image, $\bm{x}$, i.e. :
\begin{align}
  v^c_i &= \sum_n^J \sum_m^J \bigg( x^c_{nm} \max (0, 1-|z_1^T -
  m|) \max (0, 1-|z_2^T - m|) \bigg) \label{diff_sampling} \\
  v^c_i &= \sum_n^J \sum_m^J x^c_{nm} M_{nm} \approx
               \sum_n^J \sum_m^J \hat{x}^c_{nm}
          M_{nm}[M_{nm} > \epsilon] \\
  &\approx  \sum_n^J \sum_m^J \hat{x}^c_{nm}
          \ \text{nonzero}(M_{nm}[M_{nm} > \epsilon])
\end{align}

Note that $\bm{z} = [z_0, z_1, z_2] = [s, x, y]$ are the outputs of
our learned approximate posterior posterior, $q_{\bm{\phi}}(\bm{z}_i |
\bm{c}, \bm{z}_{<i})$, and correspond to the parameters of the affine transform
required by STs. The entire ST operand can be interpreted as a
multiplication (of the original image $\bm{x}$) with an element-wise
mask, $M$, with elements in the interval [0, 1]. The mask, $M$, has a
value of 1 where the $z_1$ and $z_2$ co-ordinates match the outputs of
the posterior distribution and values less than 1 outside this
zone. Taking this, coupled with the fact that many regions will have
 weight contributions near zero, we can
truncate the mask to a specific region of interest and only load
$\bm{\hat{x}}^c$, the image crop. In addition, instead of generating
the mask, $M$, in the space of the full-resolution image (which might
also be infeasible), we generate a low-resolution mask, truncate it to
the ROI and upsample it to the size of the crop $\bm{\hat{x}}^c$. This
entire process is graphically illustrated in figure \ref{trunc_st}.
Coupling the above mentioned optimizations negates the need for loading of the entire
image and mask into CPU / GPU memory and is necessary in very large
dimensional settings. While we did not
need this for our experimental setup (as the images fit into GPU
memory), we validated the method on a small experiment in Appendix
Section \ref{eval_lst}.

\subsection{Image Format and Quick Region Extraction}

Most standard image formats such as PNG and JPEG do not support random
access, i.e. $$\bm{x}[z_1:z_1+z_0J, z_2:z_2+z_0J]$$ (where $J \times J$ are the
dimensions of the image). To allow for
rapid processing of large images, we utilized Pyramid TIFF images,
coupled with the VIPS \cite{martinez2005vips} library. This choice
reduces the extraction of crops from multiple seconds to milliseconds per
image. This approach scales logarithmically with increase in image
dimensionality, allowing us to use ultra-large images. In order to
further improve performance, we also parallelized the crop-extraction for mini-batches across multiple CPU
cores.

\subsection{Evaluating the Localized Spatial Transformer}\label{eval_lst}
The localized spatial transformer (LST) from Section
\ref{local_spatial_xformer} is a necessary component when the size of
the images increases to a capacity greater than the available GPU
memory. In this case it is not possible to:

\begin{enumerate}
\item Load the image $\bm{x}$ into memory.
\item Generate the appropriately sized mask $M$ due to the same constrant.
\end{enumerate}

In order to validate that the LST works as
intended, we utilize a simple Single-Digit-Cluttered MNIST dataset
wherein the objective is simply to predict the correct digit present
in the image. We contrast the LST to a standard spatial transfomer (ST) and visualize
the test accuracy below.

\begin{figure}[H]
  \begin{center}
    \includegraphics[width=0.6\linewidth]{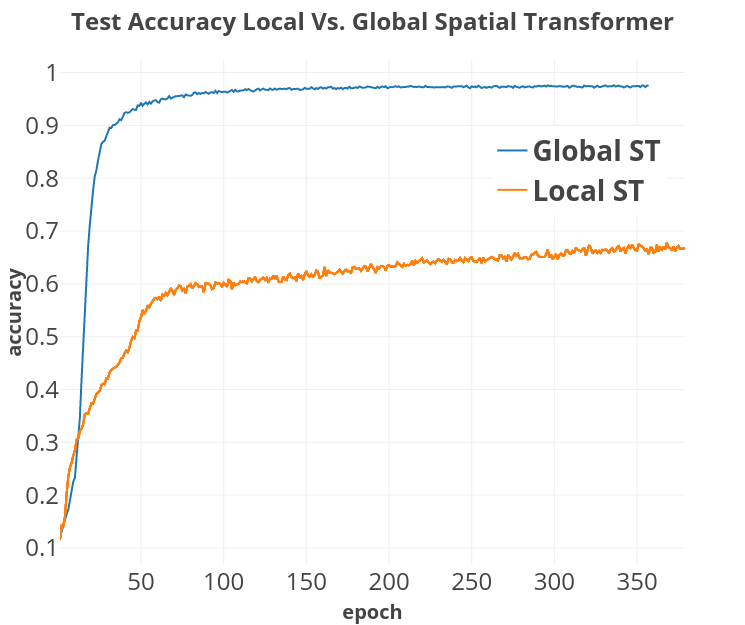}
  \end{center}
  \vskip -0.4in
\end{figure}

We observe that the LST (68\%) does not perform as well as the ST. We
believe that this is due to the fact that the full ST (97\%) has access
to more surrounding pixels which in turn provide a better gradient
estimate. In this example we only tested the case where the crop
matched the support of the affine grid. See
\textbf{models/localized\_spatial\_transformer.py} in the code-base\footnote{\url{https://github.com/jramapuram/variational_saccading}}
for more details. We plan on doing a hyper-parameter sweep by adding
extra pixels outside the support provided by the affine grid and
believe that this will improve prediction performance.


\subsection{Model}\label{model_desc}
\begin{figure}[H]
\begin{center}
  \centerline{\includegraphics[width=\linewidth]{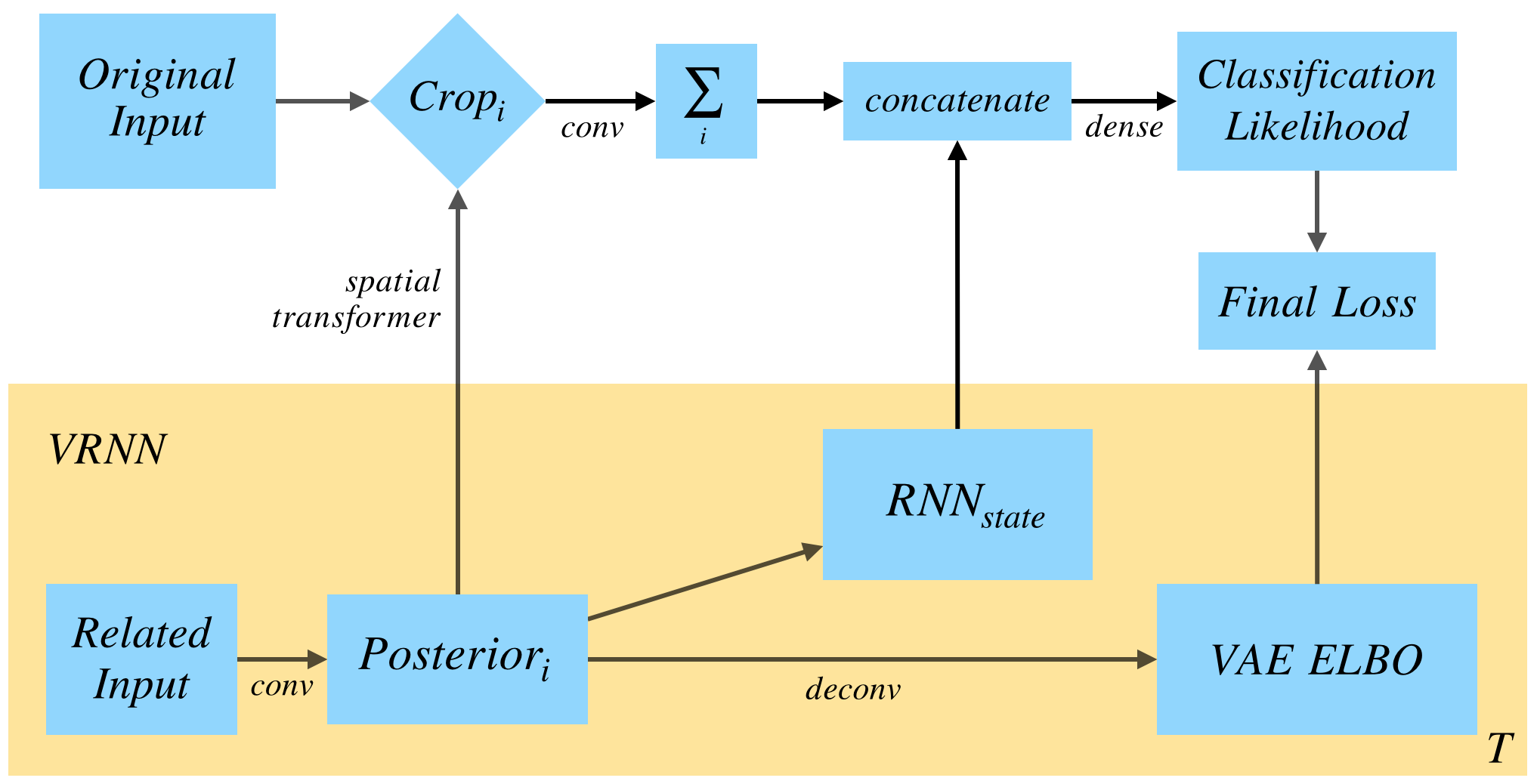}}
  \caption{Implementation of our model.} \label{model_impl}
\end{center}
\vskip -0.4in
\end{figure}

We use an isotropic Gaussian VRNN \cite{chung2015recurrent} to model our newly derived
lower bound (Equation \ref{eqn4}). The VRNN makes two crucial
modifications to the traditional ELBO:
\begin{enumerate}
\item Rather that assuming a non-informative prior, $p_{\bm{\theta}_P}(\bm{z}_i|\bm{c}, \bm{z}_{<i})$ is
learned as a function of the previous RNN hidden state, $\bm{h}_{i-1}$,
and
\item The decoder, $p_{\bm{\theta}_c}(\hat{\bm{c}} | \bm{z}_{\leq
    i})$, encoder, $q_{\bm{\phi}}(\bm{z}_i | \bm{c}, \bm{z}_{< i})$,
  and prior, $p_{\bm{\theta}_P}(\bm{z}_i | \bm{c}, \bm{z}_{<i})$, are conditioned on the
previous RNN hidden state, $\bm{h}_{i-1}$.
\end{enumerate}

\scalebox{0.90}{\parbox{1.0\linewidth}{%
\begin{align}
  \begin{split}
  \small
  p_{\bm{\theta}_p}(\bm{z}_i | \bm{z}_{<i}, \bm{c}) &= \mathcal{N}(\bm{\mu}_i(\bm{h}_{i-1};
  \bm{\theta}_{{\mu}_p}),\ \bm{\sigma}^2_i(\bm{h}_{i-1};
                  \bm{\theta}_{\sigma^2_p})) \\
  p_{\bm{\theta}}(\hat{\bm{c}}|\bm{z}_{\leq i}) &=
                    \mathcal{N}(\bm{\mu}_i(g(\bm{z}_i), \bm{h}_{i-1};
  \bm{\theta}_{\mu_c}),\ \bm{\sigma}^2_i(f_p(\bm{z}_i), \bm{h}_{i-1};
                                             \bm{\theta}_{\sigma^2_c})) \\
  q_{\bm{\phi}}(\bm{z}_i | \bm{c}, \bm{z}_{<i}) &=
  \mathcal{N}(\bm{\mu}_i(f_q(\bm{z}_i), \bm{h}_{i-1};
  \bm{\phi}_{\mu_q}),\
  \bm{\sigma}^2_i(f_q(\bm{z}_i), \bm{h}_{i-1};
  \bm{\phi}_{\sigma^2_q})))
  \end{split}\nonumber
\end{align}
}}

This dependence on $\bm{h}_{i-1}$ allows the model to integrate and relay
information about its previous saccade through to the next
timestep. The full VRNN loss function is defined as:

\scalebox{0.90}{\parbox{1.0\linewidth}{%
\begin{align}
  \begin{split}
\mathbb{E}_{q_{\bm{\phi}}(\bm{z}_{\leq T} | \bm{c}_{\leq T})} &\bigg(
  \sum_{i = 1}^T \log p_{\bm{\theta}}(\hat{\bm{c}}_i | \bm{z}_{\leq i})
  - D_{KL}(q_{\bm{\phi}}(\bm{z}_i | \bm{c}_{\leq i}, \bm{z}_{<i})
       || p_{\bm{\theta}_P}(\bm{z}_i | \bm{c}, \bm{z}_{<i}))
       \bigg)
     \end{split}
\end{align}
}}



We implement the VRNN using a
fully convolutional architecture where conv-transpose layers are used
for upsampling from the vectorized latent space. The crop classifier is implemented by a standard fully-convolutional
network, followed by a spatial pooling operation on the results of the
convolution on the crops, $\bm{o}_i$.
Adam \cite{kingma2014adam} was used as an optimizer, combined with ReLU
activations; batch-norm \cite{ioffe2015batch} was used for dense
layers and group-norm \cite{wu2018group} for convolutional
layers. For more details about specific architectural choices see
our project repository\footnote{\url{https://github.com/jramapuram/variational_saccading.git}}.

\end{document}